# Inferring hidden forcing in a biological oscillator using Kolmogorov–Arnold networks

Julian Szereszewski[1], Facundo Fainstein[1,2], Leandro E. Fernandez[1,2], Gabriel B. Mindlin[1,2]

1. Universidad de Buenos Aires, Facultad de Ciencias Exactas y Naturales, Departamento de Física, Ciudad Universitaria, 1428 Buenos Aires, Argentina.
2. CONICET - Universidad de Buenos Aires, Instituto de Física Interdisciplinaria y Aplicada (INFINA), Ciudad Universitaria, 1428 Buenos Aires, Argentina.

**abstract**

Inferring the forces that drive a dynamical system from partial observations is a fundamental challenge across physics, particularly when distinct underlying mechanisms produce similar observable dynamics. Here we show that the effective muscular forcing underlying avian respiratory dynamics can be reconstructed from measurements of air-sac pressure alone. Using an interpretable learning framework based on Kolmogorov–Arnold networks, we infer the governing equations of the system directly from data and uncover a nontrivial structure in the underlying forcing that is not apparent from the pressure signal, which instead suggests a relaxation-like oscillation. The reconstructed dynamics predict a two-phase activation pattern within each respiratory cycle, which we independently validate through electromyographic recordings of expiratory muscles. These results demonstrate that data-driven reconstruction of dynamical laws can reveal hidden physical structure and provide access to unobserved driving variables, establishing a general route to infer latent forces in partially observed dynamical systems.

## Introduction

Inferring the forces that drive a dynamical system from partial observations is a central challenge across physics [Brunton et al. 2022]. In many systems, only a limited set of variables is experimentally accessible, while the effective driving terms arise from hidden degrees of freedom. As a result, qualitatively different mechanisms can produce observables with similar temporal structure, making it difficult to identify the underlying dynamics from direct inspection of the data. Here we show that the effective forcing driving avian respiratory dynamics can be reconstructed from pressure measurements alone, revealing a nontrivial structure that is not apparent in the observed signal.

A common strategy consists of proposing a parametric form for the dynamical equations and fitting its coefficients to data [Cybenko 1989]. While successful in systems whose physical structure is well understood, this approach requires strong prior assumptions about the functional form of the equations. More recently, machine learning techniques have been employed as flexible function approximators to infer dynamical laws from data. Neural networks, for instance, can represent complex nonlinear vector fields without requiring an explicit model [Hornik et al. 1989]. However, such representations are typically difficult to interpret, as the learned dynamics are encoded in large numbers of parameters with no direct physical meaning. An alternative strategy seeks to infer explicit functional forms for the governing equations. Methods such as sparse identification of nonlinear dynamics (SINDy) reconstruct the vector field as a sparse linear combination of candidate basis functions [Brunton et al. 2016]. These approaches provide interpretable models but require specifying a library of functions a priori, which can bias the reconstruction when the true functional structure is unknown.

Recently, Kolmogorov–Arnold networks (KAN) have been proposed as a new class of neural architectures that represent multivariate functions as compositions of trainable univariate functions [Liu et al. 2024]. Because nonlinearities are learned along the edges of the network rather than fixed at the nodes, KAN models often admit transparent functional representations that can be inspected and simplified after training. This feature makes them particularly appealing for the reconstruction of dynamical systems, where interpretability of the inferred equations is essential.

In this work we apply KAN to reconstruct the dynamics of the avian respiratory system directly from experimental measurements of air-sac pressure in normally breathing birds. The respiratory apparatus of birds also plays a central role in vocal production, as pressure generated by respiratory muscles provides the driving force for sound generation in the syrinx. Here we focus on the respiratory dynamics observed during quiet breathing, where the mechanical interaction between muscular activation, air-sac compression, and airflow can be studied without the additional complexities associated with sound production. While simplified physical models of the avian respiratory system have been proposed [Fainstein et al. 2021], the structure of the effective muscular forcing that drives the respiratory cycle remains difficult to infer directly from pressure recordings.

By reconstructing the vector field underlying the pressure dynamics, we show that the structure of the forcing emerges directly from the inferred equations. Remarkably, although the pressure signal suggests a relaxation-like oscillation, the reconstructed dynamics reveal a qualitatively different organization of the underlying drive. In particular, the inferred forcing exhibits a two-phase activation structure within each respiratory cycle, a feature that is not apparent from inspection of the observable alone. We validate this prediction through independent electromyographic recordings of expiratory muscles. These results show that reconstructing governing equations from data, when combined with interpretable representations, can uncover hidden dynamical structure and provide access to latent driving variables in partially observed systems.

**Reconstruction of dynamical systems with Kolmogorov–Arnold networks**

Recent work has introduced Kolmogorov–Arnold networks (KANs) as a promising tool for extracting functional relationships from data [Liu et al. 2024]. These architectures are inspired by the Kolmogorov–Arnold representation theorem, which states that any continuous multivariate function can be expressed as a superposition of compositions of one–dimensional functions. In particular, for a function $f: [0,1]^n \to R$

$$f(x_1, x_2, \dots, x_n) = \sum_{q=1}^{2n+1} \Phi_q\left(\sum_{p=1}^{n} \phi_{qp}(x_p)\right), \quad (1)$$

where $\Phi_q$ and $\phi_{qp}$ are continuous univariate functions.

This representation suggests that the complexity of multivariate functions can be reduced to the learning of one–dimensional nonlinearities and their compositions. Kolmogorov–Arnold networks implement this idea in a neural–network architecture in which nonlinear functions are associated with the edges of the network, while nodes perform only summations (see Fig. 1).

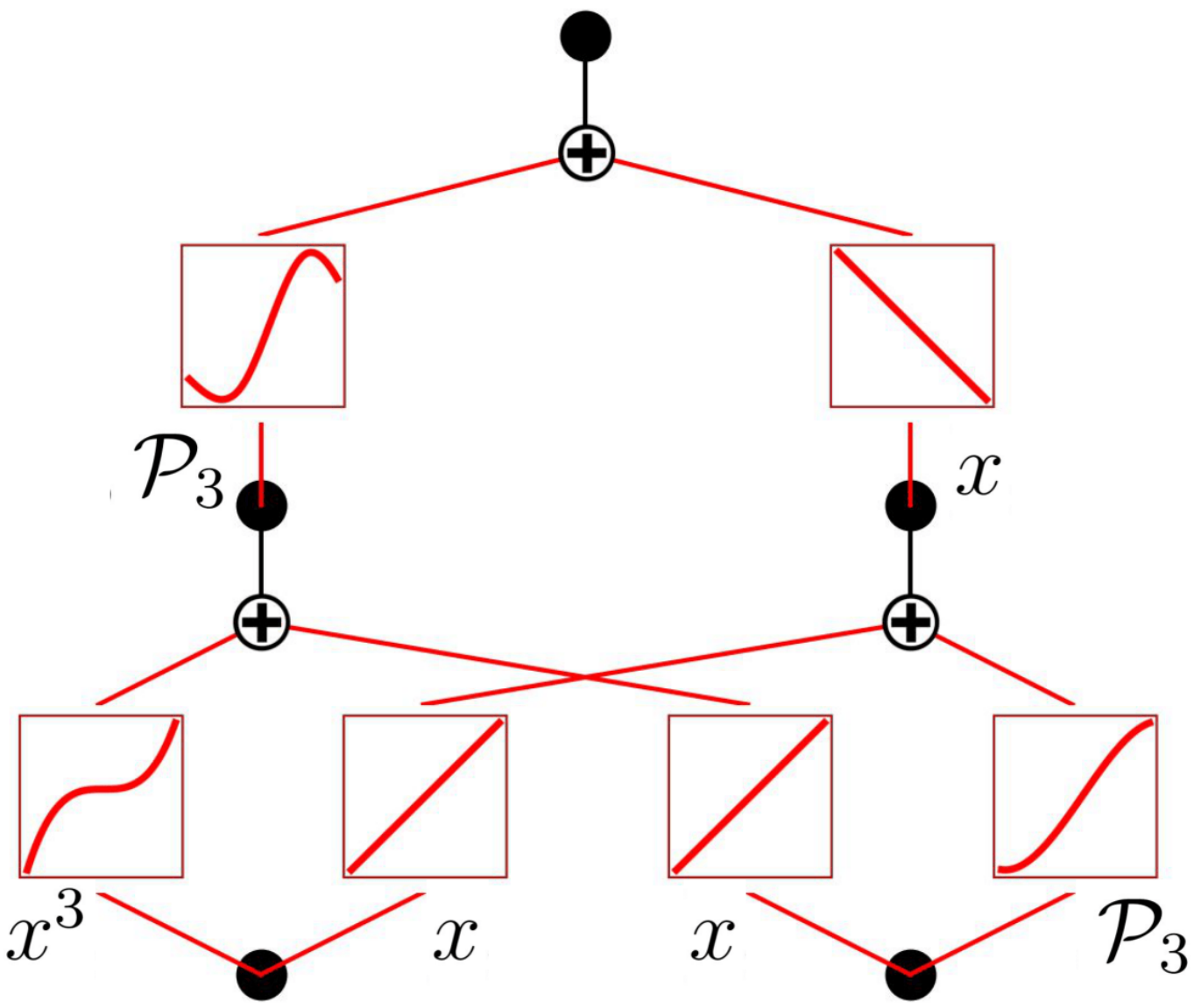

***Fig. 1 — Architecture of a Kolmogorov–Arnold network.*** *Nodes perform only summations while nonlinear transformations are associated with the edges of the network. Each edge represents a trainable one–dimensional function, typically parameterized as a spline expansion. In this example, $P_3$ denotes a vector in the space of polynomials of degree three or less. This architecture makes the functional dependencies learned by the network directly accessible for interpretation.*

More precisely, if $x_i^{(l)}$ denotes the activity of neuron $i$ in layer $l$, the next layer is computed as

$$x_j^{(l+1)} = \sum_i \phi_{j,i}^{(l)}\left(x_i^{(l)}\right), \quad (2)$$

where each $\phi_{j,i}^{(l)}$ is a trainable one–dimensional function. In practice, these functions are represented as a combination of a smooth base function and a spline expansion whose coefficients are optimized during training. This architecture differs from standard multilayer perceptrons, where nonlinearities are fixed activation functions applied at the nodes. In KANs, nonlinearities reside in the connections, making the learned functional dependence between variables directly accessible. As a result, trained networks can often be interpreted symbolically by approximating the learned univariate functions with analytic expressions.

For the purpose of dynamical-system reconstruction, the goal is to approximate the unknown vector field $\dot{x} = F(x)$ from measurements of the system's trajectory. In this work we train independent KAN models for each component of $F$, using samples of the form $(x, \dot{x})$ obtained from experimental data. Once trained, the network provides an explicit functional representation of the vector field, which can then be analyzed or simplified into symbolic expressions.

An important feature of Kolmogorov–Arnold networks is that the nonlinear dependencies learned during training remain explicitly accessible through the one–dimensional functions associated with each connection. This transparency makes it possible to inspect the functional structure of the inferred vector field rather than treating the model as a black box. In the context of dynamical systems, this provides an opportunity to identify hidden structure in the equations governing the data. In the following sections we apply this approach to experimental measurements of avian respiratory pressure. As we show below, the functions learned by the network reveal a structure of the forcing term that is not apparent from inspection of the raw pressure signal alone.

**Experimental system: avian respiratory dynamics**

The respiratory system of birds provides a well–characterized experimental system in which neural commands are translated into measurable pressure dynamics [Fainstein et al. 2021]. During vocal production, expiratory muscles compress a set of air sacs that act as reservoirs, generating the pressure that ultimately drives sound production in the syrinx. As a consequence, the temporal evolution of air–sac pressure reflects the action of respiratory muscles and provides a direct window into the motor control underlying vocal behavior. Measurements of subsyringeal pressure have therefore become a standard observable for studying the biomechanics and neural control of birdsong [Trevisan et al. 2006, Fainstein et al. 2025].

In contrast to mammalian lungs, avian lungs are relatively rigid and are ventilated by a system of anterior and posterior air sacs that expand and contract under the action of thoracic–abdominal muscles [J. N. Maina et al. 2025]. During expiration, muscular compression of the air sacs generates an increase in internal pressure, while aerodynamic valves ensure a largely unidirectional airflow through the lung. Under these conditions, the pressure inside the air sacs evolves as the result of the interaction between mechanical forcing produced by the respiratory muscles, elastic restoring forces, and viscous dissipation associated with airflow through the respiratory tract.

From the perspective of dynamical systems, the pressure signal can therefore be interpreted as the observable of a low–dimensional mechanical system driven by a time–dependent muscular forcing. Previous theoretical work has shown that the essential features of these dynamics can be captured by a reduced mechanical model describing the compression of an effective air–sac cavity coupled to airflow through the respiratory tract. In this framework, the evolution of pressure is governed by a nonlinear dynamical system driven by an effective forcing term representing the action of respiratory muscles. While such models capture the qualitative structure of the respiratory dynamics, the precise form of the effective forcing produced by the respiratory muscles is typically introduced phenomenologically. In the following sections we take a complementary approach and attempt to reconstruct the governing dynamical system directly from experimental pressure measurements using Kolmogorov–Arnold networks.

**The model for the avian respiratory system**

Let $x(t)$ denote the displacement of an effective piston representing the thoracic–abdominal wall. The motion of this piston can be modeled by a damped mechanical equation of the form

$$m\ddot{x} = -kx - \beta\dot{x} - Ap + f(t), \tag{3}$$

where $k$ is an effective elastic constant associated with the mechanical structure of the cavity, $\beta$ represents viscous dissipation, $A$ is an effective area coupling piston displacement to gauge pressure $p(t)$, and $f(t)$ represents the time–dependent forcing generated by respiratory muscles.

The pressure inside the air sac is determined by the gas contained in the cavity. Assuming an isothermal ideal gas and denoting by $V(t)$ the cavity volume, the pressure variation can be written as

$$dp(t) = \frac{N(t)k_B T}{V^2(t)} A dx - \frac{k_B T}{V(t)} dN, \tag{4}$$

where $N(t)$ is the number of particles in the cavity. Variations in pressure arise both from volume changes and from airflow through the respiratory tract. Linearizing the resulting expression around the equilibrium state leads to an evolution equation of the form

$$\dot{p} = a_1\dot{x} - a_2 p, \tag{5}$$

where $a_1$ and $a_2$ group geometric and dissipative parameters of the system.

The mechanical dynamics of the piston evolves on a faster timescale than the forcing dynamics. In this regime the piston velocity rapidly relaxes to a quasi–steady value determined by the instantaneous values of the slow variables. Under this adiabatic approximation the inertial term can be neglected and the piston velocity becomes

$$\dot{x} \simeq -\frac{k}{\beta}x - \frac{A}{\beta}p + \frac{1}{\beta}f(t). \tag{6}$$

Substituting this expression into the pressure equation yields a closed dynamical system for the variables $x$ and $p$. Eliminating $x$ leads to an effective second–order equation for the pressure of the form

$$\ddot{p} + \gamma\dot{p} + \omega_0^2 p = G(t), \tag{7}$$

where $\gamma$ and $\omega_0$ are effective parameters determined by the mechanical and flow properties of the system, and $G(t)$ is proportional to the time derivative of the muscular drive $f(t)$.

In previous modeling approaches the forcing term $G(t)$ is typically introduced phenomenologically, often approximated by simple waveforms representing the activation of respiratory muscles. In contrast, here we attempt to infer the effective dynamical equation governing the pressure directly from experimental data, allowing the functional structure of the forcing term to emerge from the data.

**Data–driven reconstruction of the respiratory dynamics**

We now seek to reconstruct the dynamical system underlying avian respiration directly from experimental measurements. The only available observable is the time series of the air sac pressure ($p$) shown in Fig. 2, recorded from naturally breathing canaries (*Serinus canaria*). Air sac pressure was measured by inserting a flexible cannula into a posterior air sac, with its free end connected to a pressure transducer (FHM-02 PGR, Fujikura, Tokyo, Japan) carried by the bird in a backpack. The signal was amplified with a custom-built variable-gain analog amplifier (up to 300× gain) and acquired at 44.15 kHz using a National Instruments data acquisition board (NI USB-6212). From the perspective of dynamical systems, this signal can be interpreted as the trajectory of a low-dimensional system driven by a time-dependent muscular forcing.

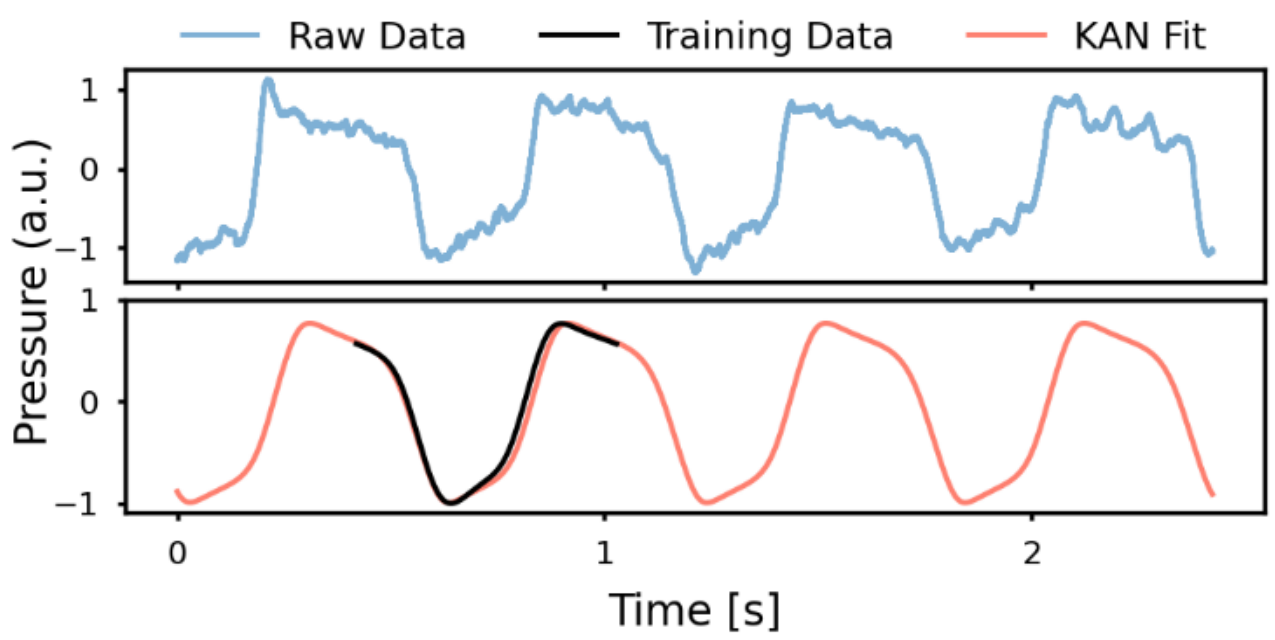

**Figure 2 — Experimental pressure signal and KAN's reconstructed dynamics.**
Top: raw air sac pressure measurements recorded during normal breathing. Bottom: training segments used for dynamical reconstruction (in black dots) together with the trajectory predicted by the fitted model (red line).

Under the reduced description introduced in the previous section and the fact that the effective dimension during quiet breathing is two, the evolution of the system can then be written in the general form

$$\ddot{p} = F(p, \dot{p}), \tag{8}$$

where the function $F$ includes the effective mechanical response of the system as well as the forcing generated by respiratory muscles.

To infer the functional form of $F$, we train a Kolmogorov–Arnold network using samples of the form $(p, \dot{p}, \ddot{p})$ obtained from the experimental pressure time series. The training data were constructed by averaging the recording over four respiratory cycles, smoothing the resulting waveform with a Savitzky–Golay filter (200-point window, first-order polynomial) and a second-order low-pass filter (8 Hz cutoff), and then computing the first and second temporal derivatives by finite differences. In this framework the network acts as a flexible functional approximator for the vector field governing the dynamics.

Unlike standard neural networks, however, the KAN architecture represents nonlinearities as explicit one–dimensional functions, which allows the learned structure of the vector field to be inspected after training. This interpretability turns out to be essential in the present context. Rather than providing only a predictive model for the pressure dynamics, the learned functions reveal a specific structure in the effective forcing driving the system. As we show below, this structure is not apparent from inspection of the pressure signal alone but becomes evident once the dynamical system is reconstructed from the data.

**Reconstruction of the vector field**

Training the Kolmogorov–Arnold network on the experimental data yields a functional representation of the vector field governing the pressure dynamics. In the reconstructed phase space $(p, \dot{p})$, the learned vector field reproduces the observed trajectories of the experimental signal and captures the oscillatory structure associated with respiratory cycles. Figure 3 shows the experimental recording together with the trajectory obtained from the inferred vector field. The reconstruction provides an explicit expression for the dynamical equation stated in Eq. 8, namely

$$\ddot{p} = c_1 p + c_2 \dot{p} + c_3 (p + c_4)^3 + c_5 (\dot{p} - c_6)^3 + c_7 (\dot{p} - c_6)^2 + c_8 (c_9 \dot{p} + (p + c_4)^3 + c_{10})^3 + c_{11} (c_9 \dot{p} + (p + c_4)^3 + c_{10})^2 + c_{12}, \tag{9}$$

where the function $F$ is represented as a composition of one–dimensional nonlinear functions learned by the network. The KAN associated to this vector field is schematized in Fig. 1.

To facilitate physical interpretation, we constrain the linear dependence on $p$ and $\dot{p}$ to match the effective mechanical response obtained in previous modeling work. Accordingly, a part of the fitted linear terms are explicitly assigned to the restoring component of the dynamics, while the remaining contributions are grouped into an effective forcing term $G = G(t)$: the temporal derivative of the muscular forcing.

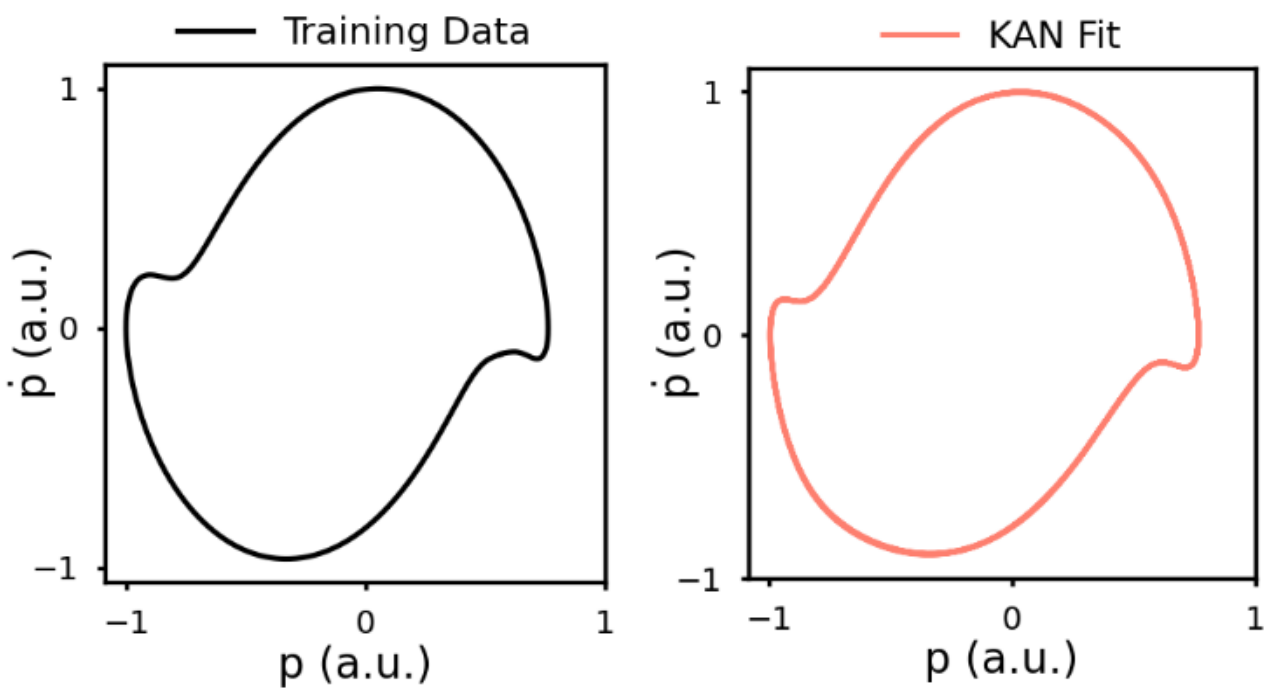

***Figure 3 — Phase–space reconstruction of the respiratory dynamics.***
*Left: experimental trajectory in the $(p, \dot{p})$ plane obtained from pressure measurements. Right: trajectory predicted by the dynamical system reconstructed with the KAN model. The reconstructed vector field reproduces the observed oscillatory cycle.*

Under this decomposition, inspection of the learned functions reveals that the dynamics can be separated into a linear restoring component depending on $p$and $\dot{p}$, and a nonlinear contribution corresponding to an effective forcing. This procedure isolates the contribution of the muscular drive from the passive mechanical response of the system.

From the reconstructed vector field, we extract the effective forcing term $G(t)$ and integrate it in time to obtain the corresponding drive $f(t)$. The inferred forcing exhibits a distinct temporal structure, characterized by two pronounced activation phases within each respiratory cycle. This feature is not apparent from inspection of the pressure signal alone, which instead suggests a simple relaxation-like dynamics [Fainstein et al. 2021].

To test this prediction, we performed independent electromyographic (EMG) recordings of expiratory muscles in naturally behaving canaries. EMG signals were measured using a pair of fine electrodes 44TDQ, Phoenix Wire Inc., VT, USA, with the exposed tips aligned in parallel and inserted into the abdominal muscles. Insulated leads were routed subcutaneously to the back of the bird and connected to a backpack, from which larger wires led to the recording unit. The signals were preprocessed with a 150 Hz high-pass RC filter and an analog differential amplifier (gain 225), both mounted on the backpack. All custom-built analog devices were powered by external 12 V batteries to minimize line noise.

Figure 4 compares the inferred forcing with the recorded EMG activity. The experimental data show strong activation both at the onset and near the end of the expiratory phase, in agreement with the structure predicted by the reconstructed dynamics. This agreement becomes even clearer when considering the EMG envelope, commonly used as a proxy for muscular force.

These features of the forcing are not apparent from the pressure recording alone. Visual inspection of the experimental time series suggests a relaxation-like oscillation driven by periodic muscular activation. However, once the dynamical system is reconstructed from the data, the functional form of the forcing becomes evident in the learned nonlinearities of the network. The KAN architecture thus provides access not only to the effective dynamics, but also to the functional structure of the underlying muscular drive. In this way, the reconstructed vector field yields a data-driven characterization of the effective forcing governing the pressure dynamics.

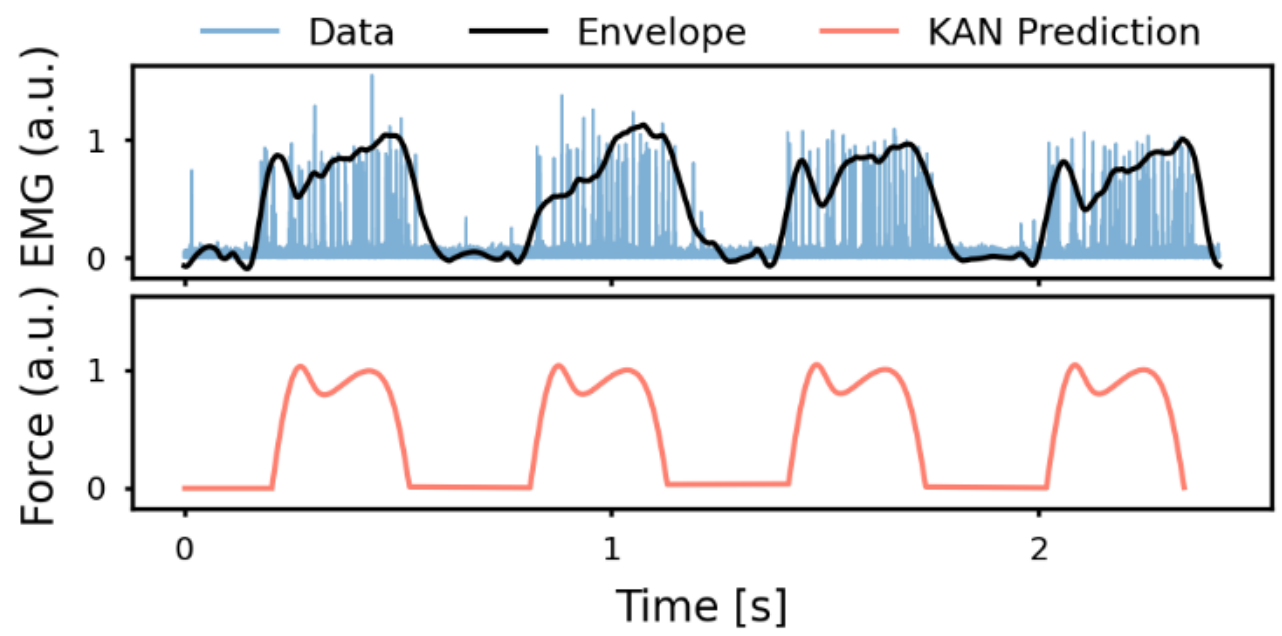

***Figure 4 — Muscular activity and inferred forcing.***
*Electromyographic (EMG) recordings of expiratory muscles together with their envelope and the forcing structure inferred from the reconstructed dynamical system. The agreement supports the interpretation that the forcing identified by the model reflects the underlying muscular drive.*

## Discussion

In this work we reconstructed the dynamical system governing avian respiratory pressure directly from experimental measurements. By combining a reduced mechanical description with a data–driven inference approach based on Kolmogorov–Arnold networks, we obtained an explicit representation of the vector field underlying the observed pressure dynamics. The reconstructed system reproduces the oscillatory respiratory cycles and provides direct access to the functional structure of the forces driving the system.

A central advantage of the KAN framework is that the nonlinear dependencies learned during training remain explicitly accessible through the one–dimensional functions associated with each connection. This transparency allows the inferred dynamics to be inspected and interpreted, rather than remaining encoded in a large set of opaque parameters as in standard neural networks. In the present case this feature proved essential: examination of the learned functions revealed a nontrivial structure in the effective forcing term driving the respiratory dynamics, a feature that is not evident from the pressure signal alone.

This approach differs from previous attempts to infer dynamical systems from data in which the functional structure of the equations must be specified in advance. For example, sparse identification methods typically rely on predefined libraries of candidate functions. In contrast, the KAN architecture allows the functional structure to emerge during training, while still yielding models that can be inspected and simplified after the fact. In this sense, the method provides a bridge between flexible machine–learning models and interpretable dynamical descriptions.

From a physical perspective, our results suggest that the effective forcing acting on the respiratory system cannot be captured by simple phenomenological waveforms. Instead, the reconstruction reveals signatures of a more complex neural control underlying the observed pressure dynamics. The reconstructed system therefore provides a compact dynamical description of the respiratory cycle that incorporates these interactions in an effective manner.

Our results show that the effective forces driving a dynamical system can be inferred from partial observations when the underlying dynamics is reconstructed in an interpretable form. In the present case, the structure of the muscular drive is not apparent from the pressure signal alone. It emerges naturally from the inferred equations and is independently validated by direct measurements. More generally, this approach provides a route to uncover hidden variables and driving mechanisms in systems where only a reduced set of observables is accessible. By combining data-driven reconstruction with physically informed interpretation, it becomes possible to extract mechanistic insight from complex biological dynamics beyond what is directly observable.